# Comparative analysis of common edge detection techniques in context of object extraction

S.K. Katiyar    P.V. Arun

Department Of Civil Engineering
MA National Institute of Technology, India

## ABSTRACT

Edges characterize boundaries and are therefore a problem of fundamental importance in image processing and particularly in automatic feature extraction. In this paper a comparative study of various edge detection techniques and band wise analysis of these algorithms in the context of object extraction with regard to remote sensing satellite images from the Indian Remote Sensing Satellite (IRS) sensors LISS-III, LISS-IV & Cartosat-I as well as Google Earth is presented. The comparison has been done between commonly used edge detection algorithms like Sobel, Canny, Prewitt, Roberts, Laplacian and Zero Crossing. Analysis results have shown that the Canny's algorithm is best suitable for the object extraction in most contexts due to fact that it yields less number of false edges, while Sobel is also a good option with lesser time and space complexity. The band wise analyses of the algorithms have also been done to find the suitability of band for the extraction of various features and it has been observed that linear features like roads, railway lines etc. can be detected more efficiently using infra red wavelength range images.

## 1. INTRODUCTION

Edges detection is a problem of fundamental importance in object extraction as it reduces image data and facilitates object detection [8] [9]. Edges identify object boundaries and are detected through abrupt changes in gray level above a particular threshold. Operators that are sensitive to the change in gray levels can be used as edge detectors. Literature reveals a great deal of edge detection techniques that can be generally classified in to three groups namely gradient, template and morphology based. Gradient based approach adopts a derivative operator to identify locations of large intensity changes. The second type resembles a template matching scheme, where the edge are modelled by a small image showing abstract properties of a perfect edge and falls under Laplacian based detection. The final type uses mathematical model of edges for detection and is a recent advancement in this context.

Sobel operator is one of the most commonly used detection methods and returns edges at points where the gradient of image intensity is maximum. Locations whose gradient value exceeds some threshold are declared edge locations [5]. Prewitt method finds edges using the Prewitt approximation to the derivative and returns edges at those points where gradient of image intensity is maximum. Prewitt operator [22] does not place any emphasis on pixels that are closer to the centre of the masks. The Canny edge detector is considered as the standard methodology of edge detection [5] and it finds edges by looking for local maxima of the gradient of Image [7]. The gradient is calculated using the derivative of a Gaussian filter and the detected edges are refined with non-maximal suppression and hysteresis. This method uses two thresholds to detect strong and weak edges, and includes the weak edges in the output only if they are connected to strong edges. This method is therefore less likely than the others to be "fooled" by noise and more likely to detect true weak edges [6] [7]. The Laplacian method searches for zero crossings in the second derivative of the image to find edges since the second derivative is zero when the first derivative is at maximum [4]. However, this method is sensitive to noise, which should be filtered out before edge detection. Based on the filter used, the two methods namely Laplacian & Gaussian and Zero crossing are more popular and these are based on Gaussian filter and specified filter respectively [12].

Expansion Matching (EXM) method [19] [13] is a template matching approach that matches a given template with a given image by expanding the image signal in terms of non orthogonal Basis Functions (BF's) which are all translated versions of the template. A template matching edge detection method based on edge expansion matching was proposed by K.R. Rao et.al [13][18-21] in which authors suggested an analytical approach which is generalized so that it easily yields the optimal SNR operator for any desired edge model. The other contemporary methods follow the approach of defining a specific set of criteria for the given edge model and optimizing them in most cases, using numerical methods. The Step Expansion Filter out performs canny in terms of DSNR but canny optimizes a combination of SNR, localization and multiple suppression criteria and is optimal in that sense [13].

The availability of different bands have enhanced various analyses however certain bands are preferable for analysis of specific features. The edge detection operator are found to be sensitive to specific bands.In this paper we analyzes most commonly used Gradient and Laplacian based edge detection techniques for various satellite images in the context of object extraction. The basic factors of concern in the context of object extraction from satellite images are false edge detection, computational complexity depending on resolution, missing of true edges and problems due to noise etc. In this paper analysis on commonly used edge detection techniques (Gradient and Laplacian based Edge Detection) has been presented for various spatial resolution remote sensing satellite images. The comparative advantages and disadvantages of one method over another have been done by visual comparison of ground features as well as field visit assisted with DGPS instrument/mobile mapping system. The band wise analysis is also conducted to find more appropriate band for evolving optimum edge detection methodology. The sensitivity of different detection techniques with reference to spatial resolution is also investigated.

## 2. DATA RESOURCES AND STUDY AREA

The investigations of present research work have been carried out for the satellite images of different spatial resolution sensors for the Bhopal city in India and details are given in the table-1. The study area is new market area of Bhopal city having area central point coordinates 23° 55' N Latitude and 76° 57' E Longitude. Various edge detection algorithms have been implemented in the MATLAB software environment. The Erdas-Imagine v9.1 was also used for the pre-processing of images and other image analysis tasks.

**Table 1:** Details of satellite images used for analysis

| S.No. | Imaging sensor | Spatial resolution(m) | Satellite | Area | Date of Image Acquisition |
|-------|----------------|----------------------|-----------|------|---------------------------|
| 1 | PAN | 2.5 | IRS-P5(Cartosat-1) | Bhopal (India) | 5th April 2009 |
| 2 | LISS-III | 24 | IRS P6 | Bhopal (India) | 5th April 2009 |
| 3 | LISS-IV | 2.5 | IRS P5 (Cartosat-1) | Bhopal (India) | 16th March 2010 |
| 4 | Google Earth | NA | NA | Bhopal (India) | 16th March 2010 |
| 5 | LANDSAT-5TM | 32.6 | LANDSAT | Bhopal (India) | 26th Dec 1990 |

### 3. METHODOLOGY AND RESULTS OF INVESTIGATIONS

The satellite images of study area have been selected in such a way that they cover some known different land use ground features with well defined and arbitrary geometry as mentioned in the investigation result tables. The important component of any edge detection method is the selection of parameters like threshold value, kernel, intensity interpolation method etc. Different edge detection techniques were analyzed using different sample images with regard to the context of object extraction. It was found that certain features can be better extracted using specific algorithms. The complexity analyses of the various edge detection methods were done using standard analysis techniques.

The parameters that we used for analysis are the ideal threshold value for the detection of the various images, the noticeable features in the edge image etc. The minimum value of threshold below which the features are not distinguishable and the maximum value of threshold above which the edges will get eliminated are determined for each of the satellite images and found that the values vary as a function of resolution as well as the edge detection techniques used. The best threshold value is also determined for each of the algorithms with reference to each satellite images. The images are subjected to various edge detection methods and a visual comparison of the results is given in the following paragraphs.

**3.1 Analysis of LISS3 sensor image**

The LISS3 sensor image was subjected to various edge detection methods and the results are given in the Figure 1. The threshold value ranges for the detection of edges formed by various ground features are also presented in the Table-2.

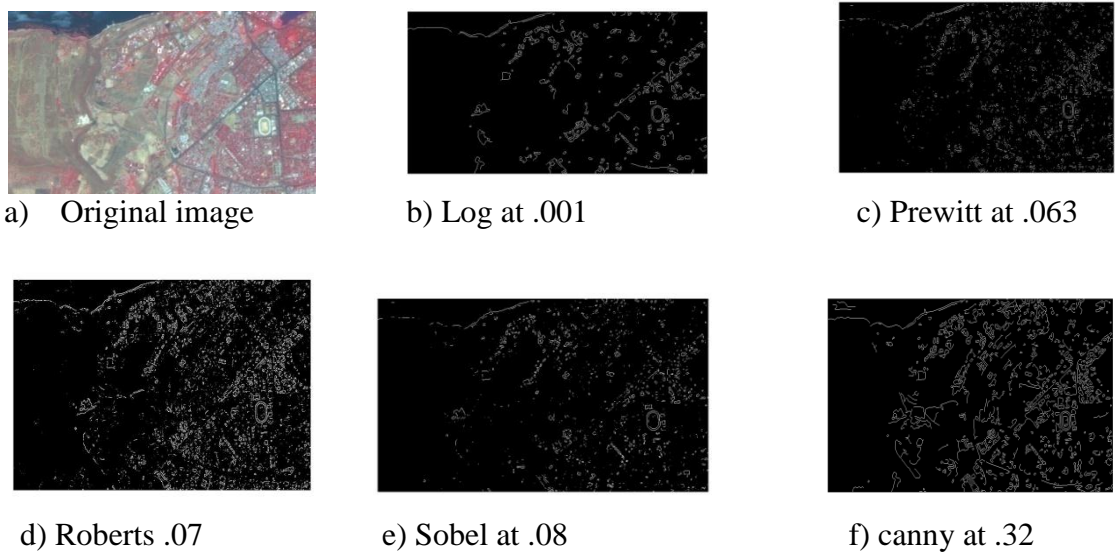

a) Original image  b) Log at .001  c) Prewitt at .063

d) Roberts .07  e) Sobel at .08  f) canny at .32

**Figure 1:** Input image and detected edges using various methods (a) to (f)

As shown in the above detected edge image as well as Table 3, more features were distinguished, when canny was used and there was not much difference in time and space complexity for the various algorithms.

**Table 2**: Analysis result of Edge Detectors for LISS3 image

| S. NO | Algorithm for edge detection | Threshold values | | | Distinguished Features on the edge image |
|---|---|---|---|---|---|
| | | min | Ideal | Max | |
| 1 | Sobel | 0 | 0.0515 | 0.3800 | Stadium, Lake |
| 2 | Canny | 0 | 0.3500 | 0.8600 | Roads, Stadium, Lake |
| 3 | Robert | 0 | 0.0533 | 0.2800 | Lake |
| 4 | Prewit | 0 | 0.0506 | 0.3400 | Lake |
| 5 | Laplacian of gausian | 0* | 0.0025 | 0.2100 | Lake |
| 6 | Zero crossing | 0* | 0.0025 | 0.2300 | Lake |

### 3.2 Analysis of LISS4 sensor Image

The LISS4 sensor image was subjected to various edge detection methods and the results are given in the Figure 2. The threshold values for the detection of edges formed by various ground features are also presented in the Table-4

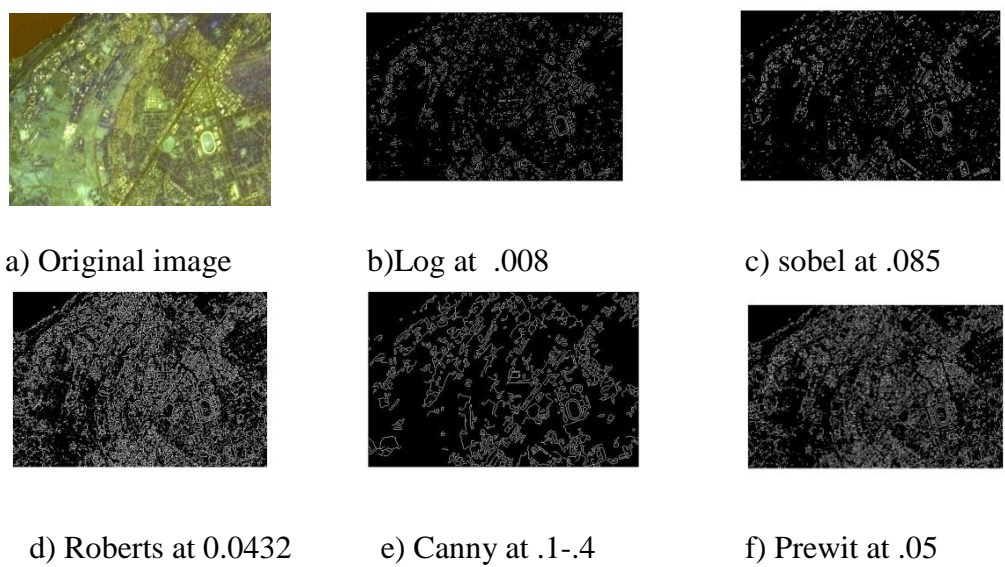

a) Original image    b) Log at .008    c) sobel at .085

d) Roberts at 0.0432    e) Canny at .1-.4    f) Prewit at .05

**Figure 2:** Input image and detected edges using various methods (a) to (f)

As shown in the Table-3 the performance of Canny and sobel are almost same and Sobel is computationally simple.

**Table 3:** Analysis result of edge detectors for LISS4 image

| S.NO | Algorithm | Threshold | | | Distinguished Features |
|---|---|---|---|---|---|
| | | Min | Ideal | Max | |
| 1 | Sobel | 0 | 0.0753 | 0.2000 | Roads, Stadium, Lake |
| 2 | Canny | 0 | 0.0250-0.0625 | 0.4750 | Roads, Stadium ,Lake |
| 3 | Robert | 0 | 0.0887 | 0.3400 | Lake |
| 4 | Prewit | 0 | 0.0736 | 0.2800 | Lake |
| 5 | Laplacian of gausian | 0* | 0.0032 | 0.3000 | Lake |
| 6 | Zero crossing | 0* | 0.0032 | 0.3000 | Lake |

The roads can be extracted to some extent using canny but breaks are there due to disturbances as shadow. The Zero crossing methods can detect the stadium areas and water bodies but the gradient methods are preferred.

### 3.3 Analysis of Google earth image

The Google Earth image was subjected to various edge detection methods and the results are given in the Figure 3. The threshold values for the detection of edges formed by various ground features are also presented in the Table-4.

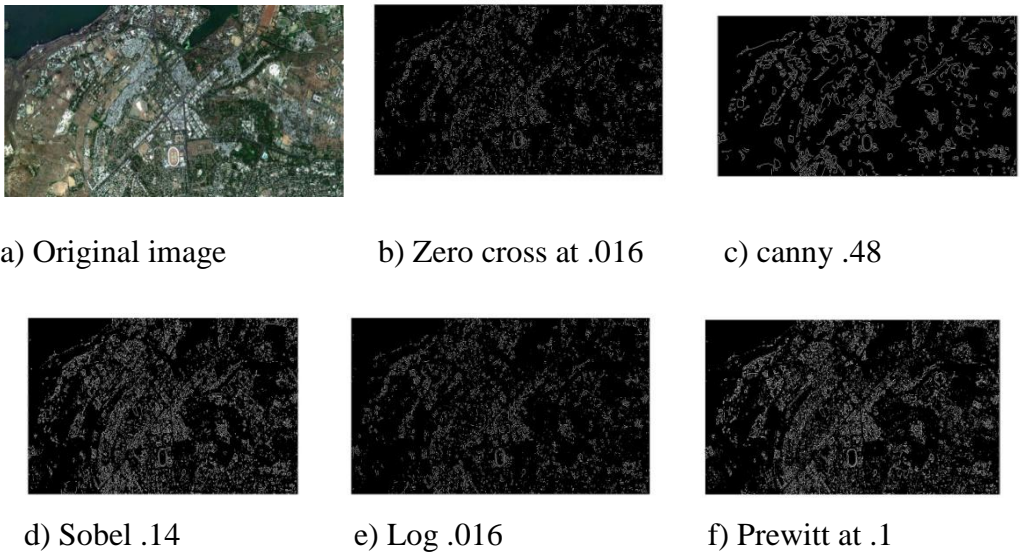

a) Original image     b) Zero cross at .016     c) canny .48

d) Sobel .14     e) Log .016     f) Prewitt at .1

**Figure 3:** Input image and detected edges using various methods (a) to (f)

The results of the analysis are summarised in the table-5 and it is evident that Canny is giving better results as compared to Sobel which is more prone to false edges. However, the computational and space complexity of canny is more when compared to others.

**Table 4:** Analysis result of edge detectors for Google earth image

| S.NO | algorithm | Threshold | | | Distinguished Features |
|---|---|---|---|---|---|
| | | Min | Ideal | Max | |
| 1 | sobel | 0 | 0.2044 | 0.5600 | Roads, Stadium |
| 2 | canny | 0 | 0.0813 0.2031 | 0.6300 | Roads, Stadium, Lake, Buildings, vehicles |
| 3 | robert | 0 | 0.2521 | 0.4100 | Lake |
| 4 | prewitt | 0 | 0.191 | 0.2300 | Lake |
| 5 | Laplacian of gausian | 0* | 0.0088 | 0.2000 | Lake |
| 6 | Zero crossing | 0* | 0.0088 | 0.2400 | Lake |

### 3.4 Analysis of PAN sensor image

The PAN sensor image was subjected to various edge detection methods and the results are given in the Figure 4. The threshold values for the detection of edges formed by various ground features are also presented in the Table-5

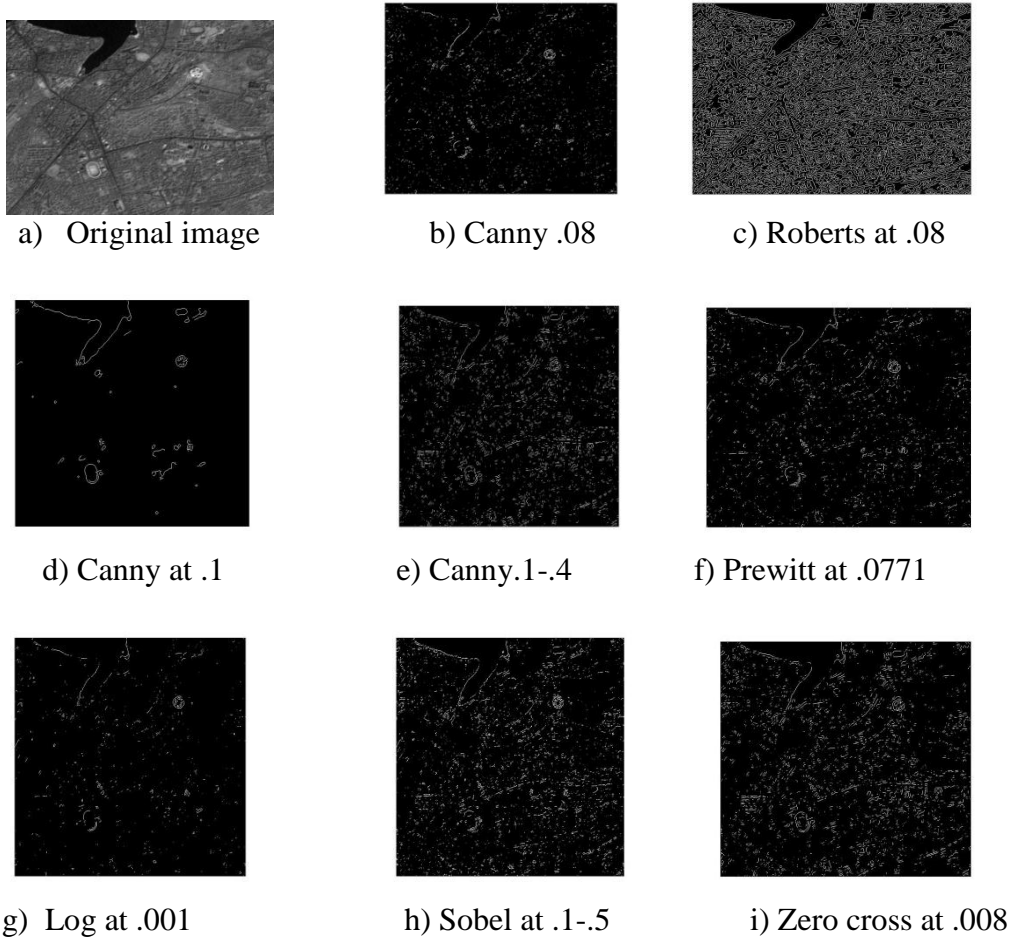

| | | |
|---|---|---|
| a) Original image | b) Canny .08 | c) Roberts at .08 |
| d) Canny at .1 | e) Canny.1-.4 | f) Prewitt at .0771 |
| g) Log at .001 | h) Sobel at .1-.5 | i) Zero cross at .008 |

**Figure 4:** Input image and detected edges using various methods (a) to (i)

The above analysis indicate that the Canny can be used to detect the detailed objects from PAN images and the features extracted, threshold values are as given

**Table 5:** Analysis result of Edge Detectors for PAN sensor image

| S.NO | algorithm | Threshold | | | Distinguished Features |
|---|---|---|---|---|---|
| | | Min | Ideal | Max | |
| 1 | sobel | 0 | 0.0797 | 0.3000 | Roads, Stadium, Lake |
| 2 | canny | 0 | 0.0563 0.1406 | 0.3500 | Roads, Stadium, Lake, Buildings |
| 3 | robert | 0 | 0.0892 | 0.2400 | Lake |
| 4 | prewit | 0 | 0.0771 | 0.3100 | Lake |
| 5 | Laplacian of gausian | 0* | 0.0043 | 0.2000 | Lake |
| 6 | Zero crossing | 0* | 0.0043 | 0.2000 | Lake |

The analysis confirms that the PAN image can be used to detect the, water bodies using canny or sobel where canny is preferred. The false edges are more at low threshold and ideal threshold causes the elimination of various weak edges.

**3.5 Analysis of CARTOSAT**

The CARTOSAT image was subjected to various edge detection methods and the results are given in the Figure 5. The threshold values for the detection of edges formed by various ground features are also presented in the Table-6

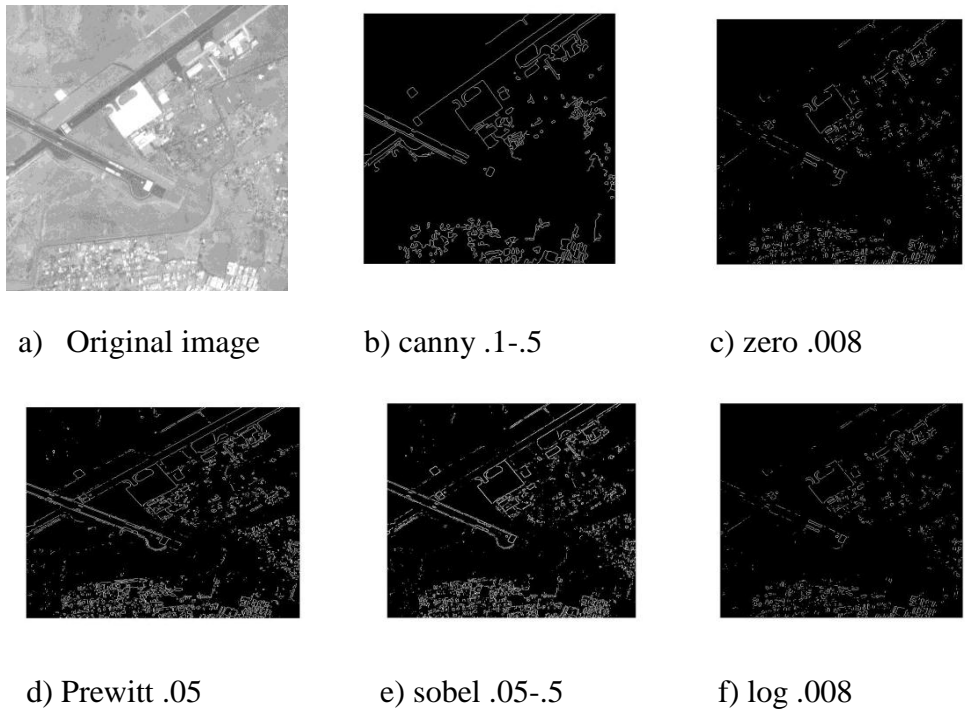

a) Original image    b) canny .1-.5    c) zero .008

d) Prewitt .05    e) sobel .05-.5    f) log .008

**Figure 5:** Input image and detected edges using various methods (a) to (f)

**Table 6:** Analysis result of Edge Detectors for CARTOSAT image

| S.NO | Algorithm | Threshold | | | Distinguished Features |
|---|---|---|---|---|---|
| | | Min | Ideal | Max | |
| 1 | Sobel | 0 | 0.0797 | 0.3000 | Roads |
| 2 | Canny | 0 | 0.0563 0.1406 | 0.4000 | Roads, Stadium, Lake, Buildings, Vehicles |
| 3 | Robert | 0 | 0.0892 | 0.1500 | Lake |
| 4 | Prewitt | 0 | 0.0771 | 0.2300 | Lake |
| 5 | Laplacian of gausian | 0* | 0.0043 | 0.2150 | Lake |
| 6 | Zero crossing | 0* | 0.0043 | 0.2340 | Lake |

The analysis confirms that the CARTOSAT image due to its high resolution is best suited for the, detection of features. Features as water bodies can be detected using canny or sobel where canny is preferred. The false edges are more at low threshold and ideal threshold causes the elimination of various weak edges

## 4. BAND WISE ANALYSIS

The images were analysed band wise also to determine the sensitivity of various edge detection algorithms to image acquisition wavelength range. Based on the basic analysis we found that the Canny & Sobel are giving better results, hence only these were used for the present investigations. The LISS-3 and LISS-4 images have been used for the analysis due to fact that only these sensors are providing multispectral images.

**4.1 Analysis of LISS 4 image bands**

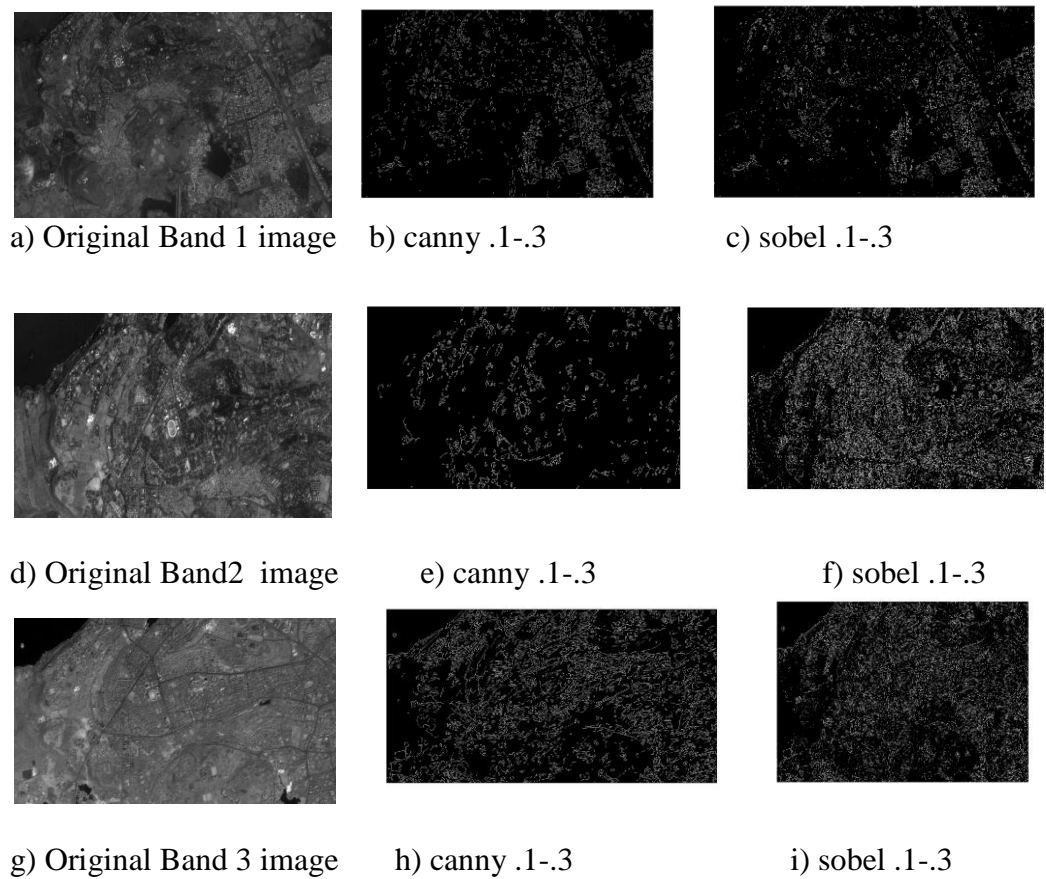

a) Original Band 1 image        b) canny .1-.3        c) sobel .1-.3

d) Original Band2 image         e) canny .1-.3        f) sobel .1-.3

g) Original Band 3 image        h) canny .1-.3        i) sobel .1-.3

**Figure 6:** Edges detected from different band LISS-3 sensor images (a) to (i)

The above analysis has confirmed that the roads can be best distinguished in band 3 and thus for extracting the roads, buildings, similar features band 3 images should be preferred.

## 4.2 Analysis of LISS 3 image bands

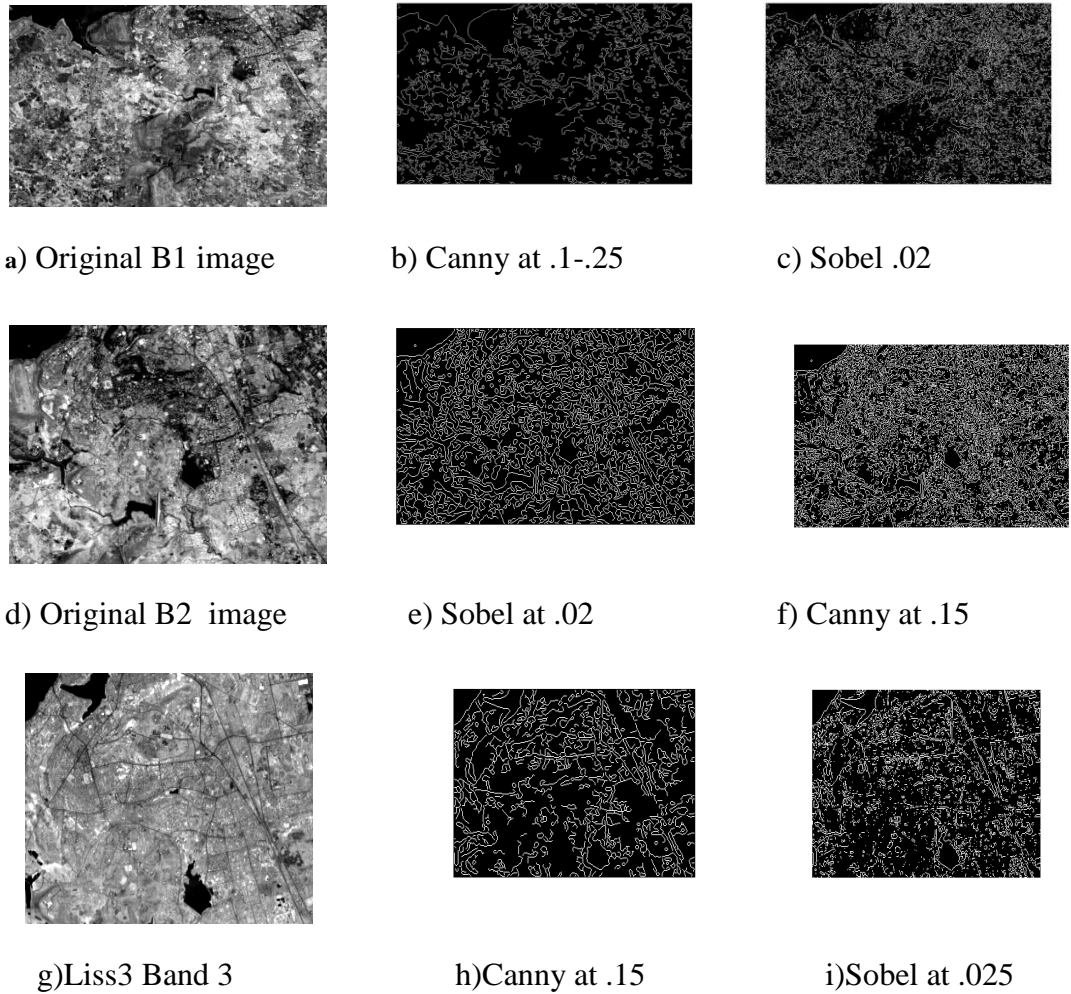

a) Original B1 image     b) Canny at .1-.25     c) Sobel .02

d) Original B2 image     e) Sobel at .02     f) Canny at .15

g) Liss3 Band 3     h) Canny at .15     i) Sobel at .025

**Figure 7:** Input image and detected edges using various methods (a) to (i)

The band wise analysis of LISS 3 images also confirms that Band3 (Infra Red Region) images are suitable for the extraction of various natural features as vegetation and also for manmade features as roads, buildings etc.

## 5. SUMMARY OF ANALYSIS

The investigations of various spatial resolution images using different edge detection methods have shown that the Laplacian based (second order) methods are more sensitive to noise than their gradient based counterparts. The comparison of each edge detection algorithm revealed that the various features can be efficiently extracted from the satellite images using Canny but Canny is computationally more complex as it takes more time as well as space. The analysis details are presented in table 7.

**Table 7:** Analysis result of Edge Detectors

| S.NO | Opertor | Complexity | | Noise sensitivity | False Edges |
|---|---|---|---|---|---|
| | | **Time** | **Space** | | |
| 1 | Sobel | lower | high | Less Sensitivity | More |
| 2 | Canny | high | high | Least Sensitivity | Least |
| 3 | Robert | high | high | Sensitivity | More |
| 4 | Prewit | low | lower | Least Sensitivity | More |
| 5 | Laplacian of Gausian | low | least | Least Sensitivity | More |
| 6 | Zero crossing | low | less | Least Sensitivity | More |

Sobel also detects the various features and is computationally more efficient as Canny but with more false edges. The other algorithms as Robert and Prewitt also detect the various features and stadium but fails in case of smaller features and the range of usable threshold is very low. The band wise analysis of the system confirms that certain bands are suitable for certain features as Infrared band for road networks, vegetation and Band2 for water bodies, buildings.

## 6. CONCLUSIONS

The investigations of present research work have led following conclusions:

1. Canny method out performs all the other methods even though its computational complexity is higher. Canny can be used for the extraction of even objects with feeble edges.
2. The Sobel also detects the various features and is computationally more efficient as Canny but with more false edges. Sobel is optimum for objects with Strong edges as lakes, Stadium etc.
3. The other algorithms as Robert and Prewitt also detect the various features and stadium but fails in case of smaller features and the range of usable threshold is very low.
4. For any method of edge detection, the computational complexity increases with the increase in the spatial resolution.
5. Also, it was found that for the extraction of certain features certain bands should be used as the response of these features to these bands are high. The IR band was found to be suitable for extraction of linear features as roads, buildings, boundaries etc. The ground truth verification confirmed that the edge detection is affected by shadows, salt and pepper noise among which the latter can be eliminated by using appropriate low pass filter.